\documentclass[10pt,journal,compsoc]{IEEEtran}

\usepackage{graphicx}
\usepackage{amsmath,amssymb}
\usepackage{booktabs}
\usepackage{tabularx}  
\usepackage[table]{xcolor}
\usepackage{pifont}
\usepackage{multirow}
\usepackage[numbers]{natbib}

\newcolumntype{Y}{>{\centering\arraybackslash}X}

\DeclareMathAlphabet{\mathcal}{OMS}{cmsy}{m}{n}
\AtBeginDocument{%
  }

\begin{document}

\title{Gram-Anchored Prompt Learning for Vision-Language Models via Second-Order Statistics}

\author{Minglei~Chen, Weilong~Wang, Jiang~Duan, and Ye~Deng%
\IEEEcompsocitemizethanks{\IEEEcompsocthanksitem All authors are with Southwestern University of Finance and Economics, Chengdu, Sichuan, China.}}

\markboth{IEEE Transactions on Pattern Analysis and Machine Intelligence,~Vol.~XX, No.~XX, Month~2026}%
{Chen \MakeLowercase{\textit{et al.}}: Gram-Anchored Prompt Learning for Vision-Language Models via Second-Order Statistics}

\IEEEtitleabstractindextext{%
\begin{abstract}
Prompt learning has become a widely used parameter-efficient strategy for adapting vision-language models (VLMs) to downstream tasks. Existing methods mainly condition text prompts on first-order visual features, such as pooled global representations or spatial tokens. Although effective for semantic discrimination, these first-order cues are often sensitive to local appearance variations and domain shifts, which limits robust adaptation. In this work, we propose \textbf{Gram-Anchored Prompt Learning (GAPL)}, a framework that introduces second-order statistical cues into prompt learning. Specifically, we construct a Gram-anchored stream that extracts compact descriptors from Gram matrices and uses them to modulate the prompted text representations. This second-order anchor complements standard first-order interactions by providing image-level structural and style information, enabling the language branch to better adapt to changes in visual statistics across domains. Extensive experiments on multiple benchmarks show that GAPL consistently improves cross-domain generalization and achieves strong overall performance.
\end{abstract}

}

\maketitle
\IEEEdisplaynontitleabstractindextext
\IEEEpeerreviewmaketitle

\IEEEraisesectionheading{\section{Introduction}}
Vision-language models (VLMs), represented by CLIP~\citep{radford2021learning}, have become a strong foundation for transferable visual recognition by learning from large-scale image-text pairs. Compared with conventional supervised models restricted to fixed label spaces, VLMs offer a flexible interface for open-vocabulary recognition and downstream adaptation. Prompt learning~\citep{zhou2022learning, zhou2022conditional, jia2022visual} has therefore emerged as a practical parameter-efficient strategy for adapting VLMs. By optimizing only a small number of prompt tokens, it can often achieve competitive performance without updating the full model.

\begin{figure}[t]
    \centering
    \includegraphics[width=\linewidth]{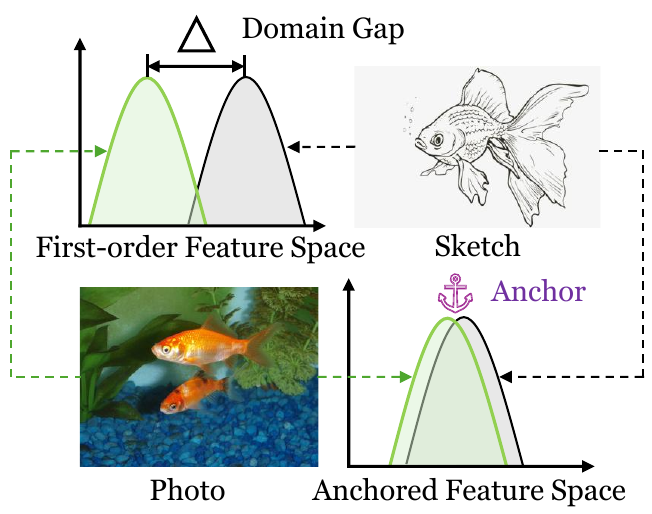}
    \caption{Illustration of first-order and anchored feature spaces. Under domain shift (e.g., photo vs.\ sketch), first-order conditioning may produce separated representations for the same semantic concept, whereas Gram-based anchoring introduces a second-order cue that brings them closer in the feature space.}
    \label{fig:teaser}
\end{figure}

However, an important challenge remains for dynamic prompt learning under domain shift. Recent methods such as CoCoOp~\citep{zhou2022conditional} improve generalization by generating image-conditioned prompts, but the conditioning signal is still mainly based on first-order visual features, such as pooled global vectors or spatial feature maps~\citep{khattak2023self, Yao_2024_CVPR}. As illustrated in Figure~\ref{fig:teaser}, these first-order features are closely tied to local appearance and spatial details. When the visual style changes from one domain to another, the same semantic concept may occupy different regions in the feature space, which makes prompt generation sensitive to style-specific variations rather than stable class structure.

This observation suggests that prompt conditioning should not rely only on first-order cues. A more robust signal can be found in a Gram-based second-order cue derived from visual features.
Prior studies in style transfer~\citep{Gatys_2016_CVPR} and domain adaptation~\citep{sun2016deep} have shown that Gram matrices capture global texture and appearance statistics that are relatively stable across style changes. This makes them a natural candidate for prompt conditioning: they can provide an image-level structural anchor that complements the semantic information from first-order features.

Based on this idea, we propose \textbf{Gram-Anchored Prompt Learning (GAPL)}, a prompt learning framework that introduces a Gram-based second-order cue into VLM adaptation.
The key component is a Gram-Anchored Stream that extracts a compact descriptor from the Gram matrix of patch tokens and uses it to modulate prompted text features. To preserve both general semantic knowledge and local discrimination, we further combine this stream with a Global Invariant Stream and a Contextual-Anchored Stream. In this way, GAPL couples global statistical alignment with local semantic alignment, leading to more stable prompt adaptation across domains.

Our main contributions are summarized as follows:
\begin{itemize}
    \item We revisit dynamic prompt learning for VLMs and identify its heavy reliance on first-order visual conditioning as an important source of sensitivity to local appearance changes and domain shifts.
    \item We propose \textbf{Gram-Anchored Prompt Learning (GAPL)}, which introduces a Gram-Anchored Stream to modulate prompted text representations with a Gram-based second-order cue derived from visual features.
    \item We build a unified framework that combines global, Gram-based, and contextual branches, and show that it improves cross-domain robustness while maintaining strong generalization on standard prompt learning benchmarks.
\end{itemize}

\section{Related Work}

\subsection{Vision-Language Models}
Vision-language models (VLMs) learn aligned image and text representations from large-scale image-text data and have become a strong foundation for transferable visual recognition. Representative models such as CLIP~\citep{radford2021learning}, ALIGN~\citep{jia2021scaling}, and LiT~\citep{zhai2022lit} map images and texts into a shared embedding space, typically through contrastive learning, and thus support open-vocabulary recognition and strong zero-shot transfer. In this work, we build on CLIP as our underlying VLM backbone. The reason is twofold. First, its visual and textual encoders offer a stable semantic alignment that is well suited for prompt-based adaptation. Second, its pre-trained feature space has been shown to retain substantial transferability even under distribution shifts, making it an appropriate testbed for studying how to improve robustness through structural prompt modulation rather than heavy parameter updating.

\subsection{Prompt Learning for VLMs}
Prompt learning has become a practical and parameter-efficient strategy for adapting VLMs to downstream tasks. Instead of updating the full backbone, prompt learning introduces a small number of learnable context tokens to steer the frozen model toward task-relevant decision boundaries. Early work such as CoOp~\citep{zhou2022learning} learns a set of static context tokens shared across all images, demonstrating that carefully optimized prompts can substantially improve few-shot performance over hand-crafted textual templates. However, because these prompts are instance-agnostic and optimized on seen categories only, they may overfit the training classes and generalize poorly to unseen categories.CoCoOp~\citep{zhou2022conditional} alleviates this issue by introducing image-conditioned prompts, which improves generalization to unseen classes. Subsequent studies further enhance prompt learning from different perspectives, including regularization against overfitting~\citep{yao2023visual, khattak2023self}, region-aware or local alignment~\citep{chenplot, lafon2024gallop}, and richer cross-modal interactions~\citep{Zheng_2025_ICCV}.

Despite these advances, most existing methods still rely primarily on first-order visual cues, such as global pooled features or spatial activations, to generate or refine prompts. Although effective for semantic discrimination, these signals can be sensitive to appearance variation and domain-specific style changes. In contrast, our method introduces a Gram-based second-order cue to guide prompt adaptation. Rather than relying solely on first-order responses, it uses a compact descriptor derived from the Gram matrix to provide a more stable anchor for cross-domain generalization.

\section{Method}
\label{sec:method}

\subsection{Overview}
\label{sec:overview}

\begin{figure*}[t]
  \centering
  \includegraphics[width=0.95\textwidth]{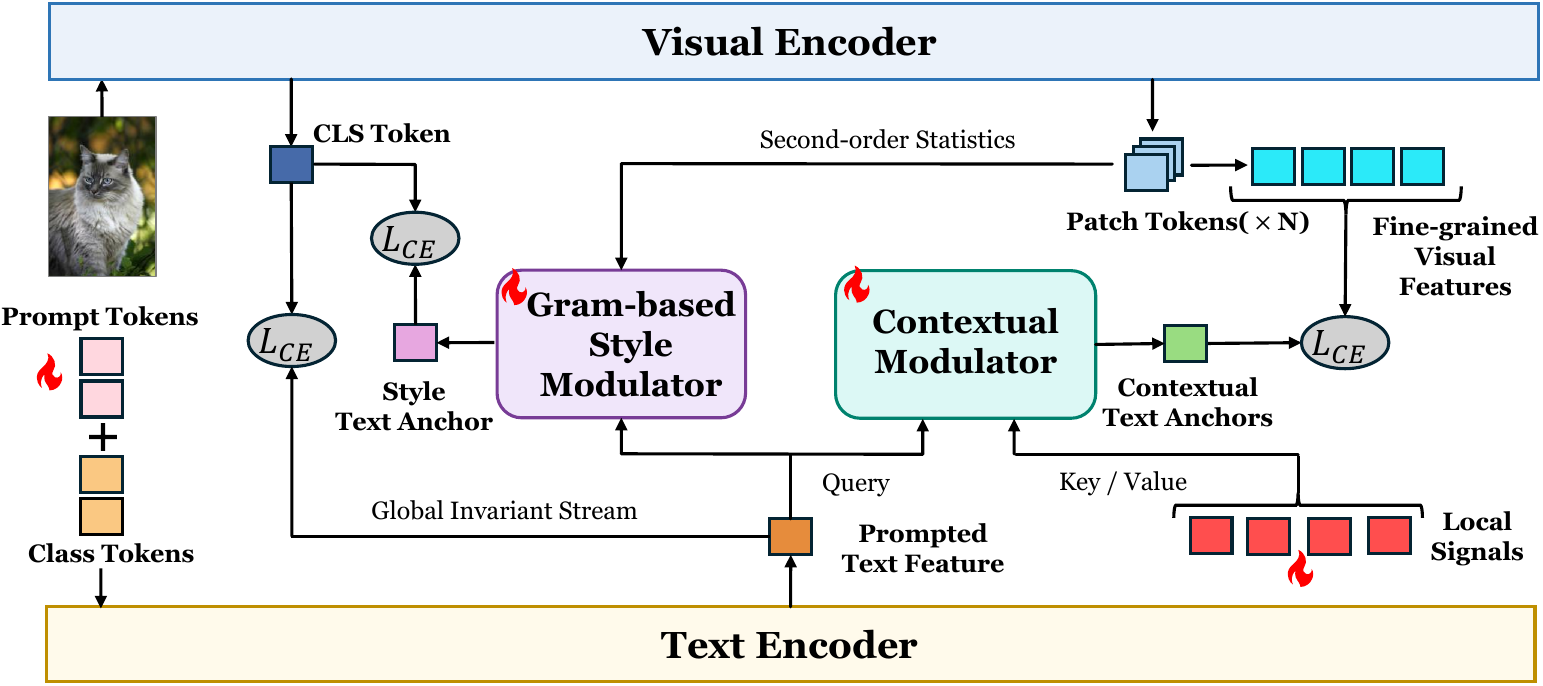}
  \caption{Overview of the proposed Gram-Anchored Prompt Learning (GAPL) framework. Learnable prompt tokens are inserted into the text encoder, and only the input-layer prompts are shown for clarity. GAPL contains three branches: (1) a Global Invariant Stream that aligns the prompted text feature with the global visual feature from the CLS token; (2) a Gram-Anchored Stream (purple), the core component of our method, which extracts a Gram-based second-order cue from patch tokens and uses a Gram-based Style Modulator to generate a Style Text Anchor; and (3) a Contextual-Anchored Stream (green), which uses learnable local signals to produce Contextual Text Anchors for fine-grained alignment. The three streams are optimized jointly for robust prompt adaptation across domains.}
  \label{fig:framework}
\end{figure*}

We propose \textbf{Gram-Anchored Prompt Learning (GAPL)}, a prompt learning framework for adapting vision-language models under domain shift. Following the deep prompting paradigm~\citep{Khattak_2023_CVPR}, GAPL inserts learnable prompt tokens into each layer of the text encoder, while Figure~\ref{fig:framework} shows only the input-layer prompts for clarity. Starting from the prompted text feature produced by the text encoder, GAPL builds three complementary streams for prediction. The first stream is a \textbf{Global Invariant Stream}, which aligns the prompted text feature with the global visual representation from the CLS token and preserves the general semantic knowledge of the pre-trained VLM. The second stream is a \textbf{Gram-Anchored Stream}, which serves as the core component of our method. 

It extracts a Gram-based second-order cue from patch tokens and feeds it into a Gram-based Style Modulator to generate a Style Text Anchor, thereby adjusting the text representation with image-level statistical information.
The third stream is a \textbf{Contextual-Anchored Stream}, which uses the prompted text feature as a query and a set of learnable local signals as keys and values to produce Contextual Text Anchors for fine-grained alignment with local visual features. These three streams play different roles: the global stream preserves stable semantic alignment, the Gram-anchored stream improves robustness to style and domain changes, and the contextual stream enhances local discrimination. They are optimized jointly and combined during inference, yielding a more robust prompt adaptation framework.

\subsection{Preliminaries}
\label{sec:preliminaries}

\noindent \textbf{Revisiting CLIP.}
CLIP~\citep{radford2021learning} aligns visual and textual representations in a shared embedding space. It consists of a visual encoder $\mathcal{V}$ and a text encoder $\mathcal{T}$. Given an input image $\mathbf{x}$, the visual encoder extracts the global feature $\mathbf{f} = \mathcal{V}(\mathbf{x}) \in \mathbb{R}^d$. For an $M$-class classification task, the text encoder generates weights $\{\mathbf{w}_c\}_{c=1}^M$ based on class-specific prompts (e.g., ``a photo of a [CLASS]''). The probability of $\mathbf{x}$ belonging to class $y$ is computed via cosine similarity:
\begin{equation}
    p(y|\mathbf{x}) = \frac{\exp(\cos(\mathbf{f}, \mathbf{w}_y)/\tau)}{\sum_{j=1}^{M} \exp(\cos(\mathbf{f}, \mathbf{w}_j)/\tau)},
\end{equation}
where $\tau$ is a temperature parameter learned by CLIP.

\noindent \textbf{Prompt Learning Setting.}
Prompt learning replaces manual context words with learnable tokens while keeping the pre-trained VLM frozen or mostly frozen~\citep{zhou2022learning,zhou2022conditional}. In this work, we follow the deep prompting paradigm~\citep{Khattak_2023_CVPR} and use the prompted text feature as the starting point for subsequent adaptation. Different from existing dynamic prompt learning methods that mainly rely on first-order image cues, GAPL augments the prompted text feature with three complementary streams: a global invariant stream, a Gram-based second-order cue stream, and a contextual stream.

\subsection{Global Invariant Stream}
\label{sec:global_prediction}

The Global Invariant Stream provides a stable semantic anchor for all subsequent branches. To reduce over-specialization to the training domain and preserve the general semantic knowledge of the pre-trained VLM, we adopt a template-based feature fusion strategy inspired by PromptSRC~\citep{khattak2023self}. Specifically, we use an ensemble of hand-crafted templates (e.g., ImageNet templates) to obtain a fixed text feature $\mathbf{w}_{fixed,c}$ for each class $c$.

During training, the model uses the learned prompted text feature $\mathbf{w}_{learn,c}$. During inference, we fuse it with the fixed template feature:
\begin{equation}
    \mathbf{w}_c=
    \begin{cases}
      \mathbf{w}_{learn,c}, & \text{during training},\\[3pt]
      \alpha \mathbf{w}_{learn,c}+(1-\alpha)\mathbf{w}_{fixed,c}, & \text{during inference},
    \end{cases}
    \label{eq:fusion}
\end{equation}
where $\alpha=0.7$ by default. This $\mathbf{w}_c$ serves both as the prediction feature of the global branch and as the base text representation for the Gram-Anchored and Contextual-Anchored streams.

The prediction of the Global Invariant Stream is
\begin{equation}
    p_{global}(y|\mathbf{x})=
    \frac{\exp(\cos(\mathbf{f},\mathbf{w}_{y})/\tau)}
    {\sum_{j=1}^{M}\exp(\cos(\mathbf{f},\mathbf{w}_{j})/\tau)}.
\end{equation}
This branch keeps the adaptation process anchored in the stable semantic space of the pre-trained VLM.

\begin{figure*}[t]
  \centering
  \includegraphics[width=0.90\textwidth]{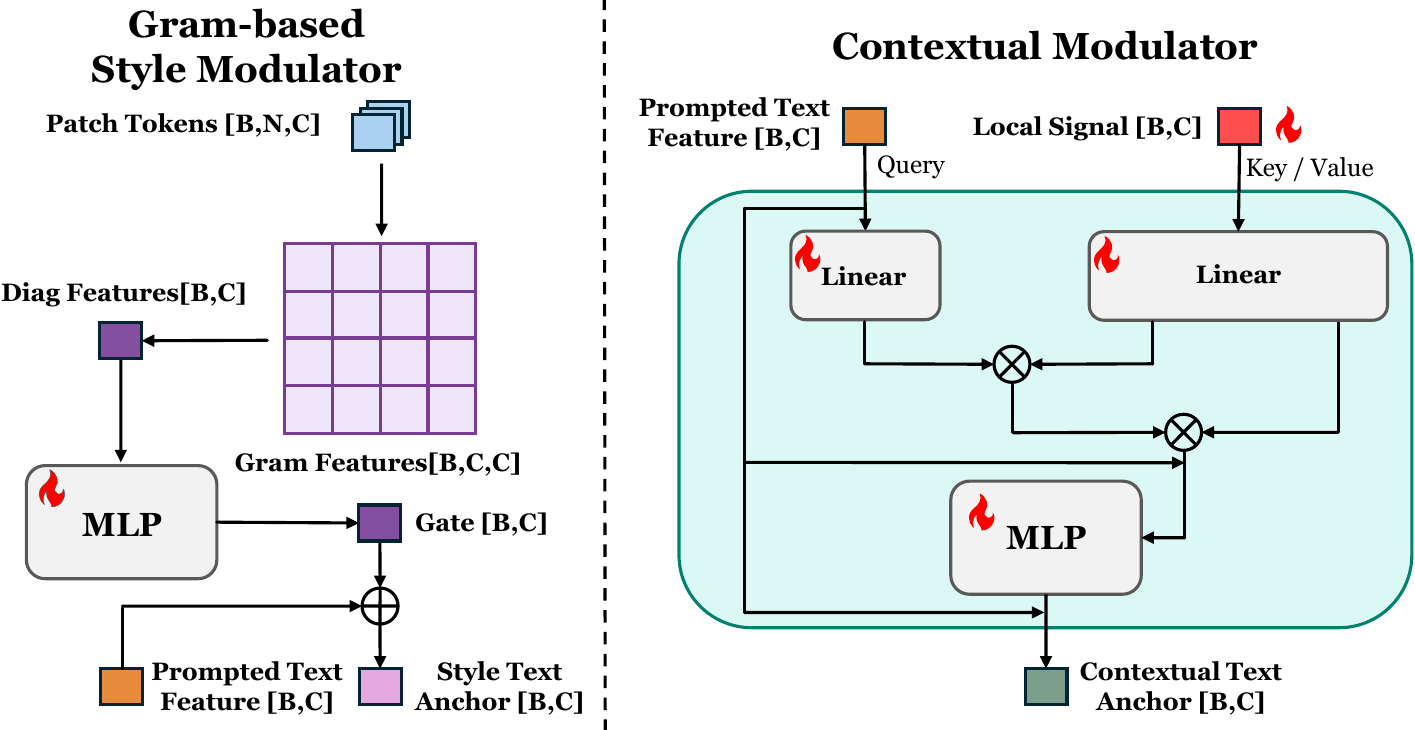}
  \caption{Detailed architecture of the Gram-based Style Modulator and the Contextual Modulator. \textbf{Left:} the Gram-based Style Modulator receives a Gram-based second-order cue derived from patch tokens, keeps their diagonal Gram features as a compact image-level descriptor, and converts them into a gating vector for modulating the prompted text feature. \textbf{Right:} the Contextual Modulator uses the prompted text feature as a query to interact with learnable local signals, producing contextual text anchors for fine-grained alignment.}
  \label{fig:modulator}
\end{figure*}

\subsection{Gram-Anchored Stream}
\label{sec:style_modulator}

The Gram-Anchored Stream is the core component of GAPL. Its goal is to introduce a Gram-based second-order cue at the image level that is less sensitive to local spatial layout than first-order features, thereby providing a more stable anchor under domain shift.

\noindent \textbf{Second-order Statistics Cue.}
Given the patch tokens $\mathbf{F}\in\mathbb{R}^{N\times d}$ from the visual encoder, we compute the channel-wise Gram matrix:
\begin{equation}
    \mathbf{G}=\frac{1}{N}\mathbf{F}^{\top}\mathbf{F}\in\mathbb{R}^{d\times d}.
\end{equation}
Rather than using the full matrix, we retain only its diagonal entries,
\begin{equation}
    \mathbf{g}_{diag}=\operatorname{diag}(\mathbf{G})\in\mathbb{R}^{d},
\end{equation}
which measure the channel-wise energy of the patch features. This yields a compact descriptor of global appearance statistics while keeping the computation simple.

\noindent \textbf{Gram-based Style Modulation.}
To convert this second-order descriptor into a text modulation signal, we first apply a log transform to compress its dynamic range, and then use a lightweight MLP to predict a gating vector:
\begin{equation}
    \mathbf{\Gamma}=
    \sigma\!\left(\Phi\!\left(\log(\mathbf{g}_{diag}+\epsilon)\right)\right),
\end{equation}
where $\sigma(\cdot)$ is the Sigmoid function and $\Phi(\cdot)$ denotes a lightweight MLP. As shown in Figure~\ref{fig:modulator}, the Style Text Anchor for class $c$ is defined as
\begin{equation}
    \mathbf{a}^{c}_{style}
    =
    \mathbf{w}_{c}+\mathbf{w}_{c}\odot\mathbf{\Gamma},
\end{equation}
where $\odot$ denotes element-wise multiplication. In this way, the prompted text feature is adjusted by a Gram-based second-order cue at the image level.

The prediction of the Gram-Anchored Stream is then computed by matching the style-modulated text anchor with the global visual feature:
\begin{equation}
    p_{gram}(y|\mathbf{x})=
    \frac{\exp(\cos(\mathbf{f},\mathbf{a}^{y}_{style})/\tau)}
    {\sum_{j=1}^{M}\exp(\cos(\mathbf{f},\mathbf{a}^{j}_{style})/\tau)}.
\end{equation}
This branch complements first-order semantic alignment with a global statistical anchor that is more robust to style and domain changes.

\definecolor{avgblue}{RGB}{244,247,255}

\begin{table*}[t]
\centering
\caption{Domain Generalization. GAPL is trained on ImageNet (16 shots) and directly evaluated on the target datasets. The best results are highlighted in \textbf{bold}.}
\label{tab:domain_generalization}
\small
\setlength{\tabcolsep}{10pt}
\setlength{\aboverulesep}{0pt}
\setlength{\belowrulesep}{0pt}
\setlength{\cmidrulesep}{0pt}
\begin{tabular}{l c cccc >{\columncolor{avgblue}}c}
    \toprule
 \multirow{2}{*}{Method} & Source & \multicolumn{4}{c}{Target (OOD Variants)} & \multicolumn{1}{>{\columncolor{avgblue}}c}{\textbf{Average}} \\
    \cmidrule(lr){2-2} \cmidrule(lr){3-6}
    & ImageNet & -V2 & -Sketch & -A & -R & \\
    \midrule
    CoOp \citep{zhou2022learning}        & 71.51 & 64.20 & 47.99 & 49.71 & 75.21 & 59.28 \\
    CoCoOp \citep{zhou2022conditional}   & 71.02 & 64.07 & 48.75 & 50.63 & 76.18 & 59.91 \\
    KgCoOp \citep{yao2023visual}         & 70.66 & 64.10 & 48.97 & 50.69 & 76.70 & 60.12 \\
    MaPLe \citep{Khattak_2023_CVPR}      & 70.72 & 64.07 & 49.15 & 50.90 & 76.98 & 60.27 \\
    TCP \citep{Yao_2024_CVPR}               & 71.20 & 64.60 & 49.50 & 51.20 & 76.73 & 60.51 \\
    MMA \citep{seputis2024multi}         & 71.00 & 64.33 & 49.13 & 51.12 & 77.32 & 60.48 \\
    CoPrompt \citep{roy2023consistency}  & 70.80 & 64.25 & 49.43 & 50.50 & 77.51 & 60.42 \\
    HiCroPL \citep{Zheng_2025_ICCV}      & 71.22 & 64.33 & 49.47 & 50.79 & 77.15 & 60.44 \\
    MMRL \citep{guo2025mmrl}             & 72.03 & 64.47 & 49.17 & 51.20 & 77.53 & 60.60 \\
    PromptSRC \citep{khattak2023self}    & 71.27 & 64.35 & 49.55 & 50.90 & \textbf{77.80} & 60.65 \\
    \midrule
    \rowcolor{avgblue}
    \textbf{GAPL (Ours)} & \textbf{72.90} & \textbf{66.43} & \textbf{50.13} & 50.43 & 77.50 & \textbf{61.12} \\
    \bottomrule
\end{tabular}
\end{table*}

\subsection{Contextual-Anchored Stream}
\label{sec:ctx_branch}

While the Gram-Anchored Stream provides an image-level statistical anchor, the Contextual-Anchored Stream focuses on local discriminative regions. It complements the global branch by improving fine-grained alignment between text features and patch-level visual content~\citep{sun2022dualcoop,zhou2022extract,lafon2024gallop}.

Given the patch tokens $\mathbf{F}\in\mathbb{R}^{N\times d}$ and a set of learnable local signals $\{\mathbf{s}_{local}^{k}\}_{k=1}^{K}$ with $K=4$, we use the Contextual Modulator to refine the prompted text feature $\mathbf{w}_{c}$ into a set of contextual text anchors:
\begin{equation}
    \mathbf{a}_{ctx}^{c,k}
    =
    \mathbf{w}_{c}
    +
    \operatorname{MLP}\!\left(
    \operatorname{Attn}(\text{Q}=\mathbf{w}_{c},\text{K}/\text{V}=\mathbf{s}_{local}^{k})
    \right).
\end{equation}
Here, $\text{Q}$, $\text{K}$, and $\text{V}$ denote the query, key, and value in the attention operator.

For each anchor $\mathbf{a}_{ctx}^{c,k}$, we compute cosine similarities with all patch tokens in $\mathbf{F}$, producing a score map $\{s_{k,n}^{c}\}_{n=1}^{N}$. To focus on the most informative regions, we sort these scores in descending order and apply hierarchical Top-$K$ pooling:
\begin{equation}
    \bar{S}_{k}^{c}
    =
    \frac{1}{M_{k}}
    \sum_{j=1}^{M_{k}}\hat{s}_{k,j}^{c},
\end{equation}
where $\hat{s}_{k,j}^{c}$ is the $j$-th largest score in the sorted list and $M_{k}=k\times 10$ is the pooling size of the $k$-th branch.

We then average the predictions across the $K=4$ contextual branches:
\begin{equation}
    p_{ctx}(y|\mathbf{x})
    =
    \frac{1}{4}
    \sum_{k=1}^{4}
    \left(
    \frac{\exp(\bar{S}_{k}^{y}/\tau)}
    {\sum_{j=1}^{M}\exp(\bar{S}_{k}^{j}/\tau)}
    \right).
\end{equation}
This branch improves local discrimination and complements the global statistical cue introduced by the Gram-Anchored Stream.

\subsection{Optimization and Inference}
\label{sec:optimization}

We train GAPL with branch-wise supervision and adaptive fusion. Let $\mathcal{M}=\{\text{global},\text{gram},\text{ctx}\}$ denote the three streams. Each stream $m\in\mathcal{M}$ produces logits $z_{m}(\mathbf{x})$ and a prediction probability $p_{m}(y|\mathbf{x})$.

To combine the three streams, we adopt an input-adaptive branch fusion strategy instead of uniform averaging. We first construct a fusion descriptor
\begin{equation}
    \mathbf{s}(\mathbf{x})
    =
    \left[\hat{\mathbf{f}}(\mathbf{x});\hat{\mathbf{\Gamma}}(\mathbf{x})\right],
\end{equation}
where $\hat{\mathbf{f}}(\mathbf{x})$ and $\hat{\mathbf{\Gamma}}(\mathbf{x})$ denote the normalized global visual feature and normalized style gate, respectively. Based on this descriptor, a two-layer MLP predicts the branch weights:
\begin{equation}
    [w_{global}(\mathbf{x}),w_{gram}(\mathbf{x}),w_{ctx}(\mathbf{x})]
    =
    \operatorname{softmax}\!\left(\Phi_{bw}(\mathbf{s}(\mathbf{x}))/T_{bw}\right),
\end{equation}
where $\Phi_{bw}(\cdot)$ is a two-layer MLP with ReLU activation and $T_{bw}$ is the temperature for branch-weight prediction.

The fused logits are defined as
\begin{equation}
    z_{fused}(\mathbf{x})
    =
    \sum_{m\in\mathcal{M}}
    w_{m}(\mathbf{x})\frac{z_{m}(\mathbf{x})}{T_{m}},
\end{equation}
where $T_{m}$ is the branch-specific temperature for stream $m$. The fused prediction is
\begin{equation}
    p_{fused}(y|\mathbf{x})=\operatorname{softmax}(z_{fused}(\mathbf{x})).
\end{equation}

The branch-wise classification loss is
\begin{equation}
    \mathcal{L}_{cls}
    =
    \sum_{m\in\mathcal{M}}\mathcal{L}_{ce}(p_{m},y_{gt}),
\end{equation}
where $y_{gt}$ is the ground-truth label. Following prior prompt learning practice, we further use regularization terms in the text and image spaces to reduce knowledge forgetting. The final training objective is
\begin{equation}
    \mathcal{L}_{total}
    =
    \mathcal{L}_{cls}
    +
    \lambda_{fused}\mathcal{L}_{ce}(p_{fused},y_{gt})
    +
    \lambda_{txt}\mathcal{L}_{reg}^{txt}
    +
    \lambda_{img}\mathcal{L}_{reg}^{img}.
\end{equation}

During inference, we use the fused prediction $p_{fused}$ as the final output.
\section{Experiments}
\label{sec:experiments}

\subsection{Datasets} Following the standard evaluation protocols in recent prompt learning literature, we conduct extensive experiments on a comprehensive benchmark comprising 15 publicly available datasets. For generalization capability analysis, we utilize ImageNet as the source domain. The downstream tasks include ImageNet~\citep{deng2009imagenet}, Caltech101~\citep{fei2004learning}, OxfordPets~\citep{parkhi2012cats}, StanfordCars~\citep{krause20133d}, Flowers102~\citep{nilsback2008automated}, Food101~\citep{bossard2014food}, FGVCAircraft~\citep{maji2013fine}, SUN397~\citep{xiao2010sun}, DTD~\citep{cimpoi2014describing}, EuroSAT~\citep{helber2019eurosat}, and UCF101~\citep{soomro2012ucf101}. Furthermore, to evaluate robustness against domain shifts, we employ four ImageNet variants: ImageNet-V2~\citep{recht2019imagenet}, ImageNet-Sketch~\citep{wang2019learning}, ImageNet-A~\citep{hendrycks2021natural}, and ImageNet-R~\citep{hendrycks2021many}.These datasets cover diverse visual recognition scenarios, including fine-grained categorization, scene understanding, texture recognition, action recognition, and remote sensing classification.

\definecolor{avgblueDeep}{RGB}{225,237,247}

\begin{table*}[t]
\centering
\caption{Base-to-novel generalization results on 11 datasets. We particularly emphasize the \emph{Average} results, as they offer a more comprehensive indicator of the balance between base-class fitting and novel-class generalization across diverse benchmarks. In particular, GAPL establishes a new state of the art in average HM, mainly driven by its improved performance on novel classes without sacrificing strong base-class accuracy. The best results are highlighted in \textbf{bold}.}
\label{tab:base_to_new}
\small
\setlength{\tabcolsep}{10pt}

\setlength{\aboverulesep}{0pt}
\setlength{\belowrulesep}{0pt}
\setlength{\cmidrulesep}{0pt}

\resizebox{\textwidth}{!}{%
\begin{tabular}{l ccc ccc ccc ccc}
\toprule
\multirow{2}{*}{Method}
& \multicolumn{3}{>{\columncolor{avgblue}}c}{\textbf{Average Performance}}
& \multicolumn{3}{c}{ImageNet}
& \multicolumn{3}{c}{Caltech101}
& \multicolumn{3}{c}{OxfordPets} \\
\cmidrule(lr){2-4} \cmidrule(lr){5-7} \cmidrule(lr){8-10} \cmidrule(lr){11-13}
& \cellcolor{avgblue}Base & \cellcolor{avgblueDeep}Novel & \cellcolor{avgblueDeep}HM
& Base & Novel & HM
& Base & Novel & HM
& Base & Novel & HM \\
\midrule
CoOp \citep{zhou2022learning}
    & \cellcolor{avgblue}82.69 & \cellcolor{avgblueDeep}63.22 & \cellcolor{avgblueDeep}71.66
    & 76.47 & 67.88 & 71.92 & 98.00 & 89.81 & 93.73 & 93.67 & 95.29 & 94.47 \\
CoCoOp \citep{zhou2022conditional}
    & \cellcolor{avgblue}80.47 & \cellcolor{avgblueDeep}71.69 & \cellcolor{avgblueDeep}75.83
    & 75.98 & 70.43 & 73.10 & 97.96 & 93.81 & 95.84 & 95.20 & 97.69 & 96.43 \\
KgCoOp \citep{yao2023visual}
    & \cellcolor{avgblue}80.73 & \cellcolor{avgblueDeep}73.60 & \cellcolor{avgblueDeep}77.00
    & 75.83 & 69.96 & 72.78 & 97.72 & 94.39 & 96.03 & 94.65 & 97.76 & 96.18 \\
MaPLe \citep{Khattak_2023_CVPR}
    & \cellcolor{avgblue}82.28 & \cellcolor{avgblueDeep}75.14 & \cellcolor{avgblueDeep}78.55
    & 76.66 & 70.54 & 73.47 & 97.74 & 94.36 & 96.02 & 95.43 & 97.76 & 96.58 \\
TCP \citep{Yao_2024_CVPR}
    & \cellcolor{avgblue}84.13 & \cellcolor{avgblueDeep}75.36 & \cellcolor{avgblueDeep}79.50
    & 77.27 & 69.87 & 73.38 & 98.23 & 94.67 & 96.42 & 94.67 & 97.20 & 95.92 \\
MMA \citep{seputis2024multi}
    & \cellcolor{avgblue}83.20 & \cellcolor{avgblueDeep}76.80 & \cellcolor{avgblueDeep}79.87
    & 77.31 & 71.00 & 74.02 & 98.40 & 94.00 & 96.15 & 95.40 & 98.07 & 96.72 \\
CoPrompt \citep{roy2023consistency}
    & \cellcolor{avgblue}84.00 & \cellcolor{avgblueDeep}77.23 & \cellcolor{avgblueDeep}80.47
    & 77.67 & 71.27 & 74.33 & 98.27 & 94.90 & 96.56 & 95.67 & \textbf{98.10} & 96.87 \\
HiCroPL \citep{Zheng_2025_ICCV}
    & \cellcolor{avgblue}\textbf{85.89} & \cellcolor{avgblueDeep}77.99 & \cellcolor{avgblueDeep}81.75
    & \textbf{78.07} & \textbf{71.72} & \textbf{74.76} & 98.77 & \textbf{95.96} & 97.34 & \textbf{96.28} & 97.76 & \textbf{97.01} \\
MMRL \citep{guo2025mmrl}
    & \cellcolor{avgblue}85.68 & \cellcolor{avgblueDeep}77.16 & \cellcolor{avgblueDeep}81.20
    & 77.90 & 71.30 & 74.45 & \textbf{98.97} & 94.50 & 96.68 & 95.90 & 97.60 & 96.74 \\
PromptSRC \citep{khattak2023self}
    & \cellcolor{avgblue}84.26 & \cellcolor{avgblueDeep}76.10 & \cellcolor{avgblueDeep}79.97
    & 77.60 & 70.73 & 74.01 & 98.10 & 94.03 & 96.02 & 95.33 & 97.30 & 96.30 \\
\midrule
\rowcolor{avgblue}
\textbf{GAPL (Ours)}
    & \cellcolor{avgblue}85.78 & \cellcolor{avgblue}\textbf{78.12} & \cellcolor{avgblue}\textbf{81.77}
    & 77.30 & 69.87 & 73.40 & 98.90 & 95.87 & \textbf{97.36} & 95.73 & 97.90 & 96.80 \\
\bottomrule
\end{tabular}%
}

\vspace{0.5em}

\resizebox{\textwidth}{!}{%
\begin{tabular}{l ccc ccc ccc ccc}
\toprule
\multirow{2}{*}{Method} & \multicolumn{3}{c}{StanfordCars} & \multicolumn{3}{c}{Flowers102} & \multicolumn{3}{c}{Food101} & \multicolumn{3}{c}{FGVCAircraft} \\
\cmidrule(lr){2-4} \cmidrule(lr){5-7} \cmidrule(lr){8-10} \cmidrule(lr){11-13}
& Base & Novel & HM & Base & Novel & HM & Base & Novel & HM & Base & Novel & HM \\
\midrule
CoOp \citep{zhou2022learning}
    & 78.12 & 60.40 & 68.13 & 97.60 & 59.67 & 74.06 & 88.33 & 82.26 & 85.19 & 40.44 & 22.30 & 28.75 \\
CoCoOp \citep{zhou2022conditional}
    & 70.49 & 73.59 & 72.01 & 94.87 & 71.75 & 81.71 & 90.70 & 91.29 & 90.99 & 33.41 & 23.71 & 27.74 \\
KgCoOp \citep{yao2023visual}
    & 71.76 & 75.04 & 73.36 & 95.00 & 74.73 & 83.65 & 90.50 & 91.70 & 91.10 & 36.21 & 33.55 & 34.83 \\
MaPLe \citep{Khattak_2023_CVPR}
    & 72.94 & 74.00 & 73.47 & 95.92 & 72.46 & 82.56 & 90.71 & 92.05 & 91.38 & 37.44 & 35.61 & 36.50 \\
TCP \citep{Yao_2024_CVPR}
    & 80.80 & 74.13 & 77.32 & 97.73 & 75.57 & 85.23 & 90.57 & 91.37 & 90.97 & 41.97 & 34.43 & 37.83 \\
MMA \citep{seputis2024multi}
    & 78.50 & 73.10 & 75.70 & 97.77 & 75.93 & 85.48 & 90.13 & 91.30 & 90.71 & 40.57 & 36.33 & 38.33 \\
CoPrompt \citep{roy2023consistency}
    & 76.97 & 74.40 & 75.66 & 97.27 & 76.60 & 85.71 & 90.73 & \textbf{92.07} & 91.40 & 40.20 & 39.33 & 39.76 \\
HiCroPL \citep{Zheng_2025_ICCV}
    & 81.51 & 75.04 & 78.14 & 98.29 & 75.46 & 85.38 & 90.96 & 91.67 & 91.31 & \textbf{48.38} & \textbf{41.75} & \textbf{44.82} \\
MMRL \citep{guo2025mmrl}
    & 81.30 & \textbf{75.07} & 78.06 & \textbf{98.97} & 77.27 & 86.78 & 90.57 & 91.50 & 91.03 & 46.30 & 37.03 & 41.15 \\
PromptSRC \citep{khattak2023self}
    & 78.27 & 74.97 & 76.58 & 98.07 & 76.50 & 85.95 & 90.67 & 91.53 & 91.10 & 42.73 & 37.87 & 40.15 \\
\midrule
\rowcolor{avgblue}
\textbf{GAPL (Ours)}
    & \textbf{84.70} & 74.50 & \textbf{79.27} & 98.63 & \textbf{77.70} & \textbf{86.92} & \textbf{90.97} & 92.00 & \textbf{91.48} & 47.80 & 41.30 & 44.31 \\
\bottomrule
\end{tabular}%
}

\vspace{0.5em}

\resizebox{\textwidth}{!}{%
\begin{tabular}{l ccc ccc ccc ccc}
\toprule
\multirow{2}{*}{Method} & \multicolumn{3}{c}{SUN397} & \multicolumn{3}{c}{DTD} & \multicolumn{3}{c}{EuroSAT} & \multicolumn{3}{c}{UCF101} \\
\cmidrule(lr){2-4} \cmidrule(lr){5-7} \cmidrule(lr){8-10} \cmidrule(lr){11-13}
& Base & Novel & HM & Base & Novel & HM & Base & Novel & HM & Base & Novel & HM \\
\midrule
CoOp \citep{zhou2022learning}
    & 80.60 & 65.89 & 72.51 & 79.44 & 41.18 & 54.24 & 92.19 & 54.74 & 68.69 & 84.69 & 56.05 & 67.46 \\
CoCoOp \citep{zhou2022conditional}
    & 79.74 & 76.86 & 78.27 & 77.01 & 56.00 & 64.85 & 87.49 & 60.04 & 71.21 & 82.33 & 73.45 & 77.64 \\
KgCoOp \citep{yao2023visual}
    & 80.29 & 76.53 & 78.36 & 77.55 & 54.99 & 64.35 & 85.64 & 64.34 & 73.48 & 82.89 & 76.67 & 79.66 \\
MaPLe \citep{Khattak_2023_CVPR}
    & 80.82 & 78.70 & 79.75 & 80.36 & 59.18 & 68.16 & 94.07 & 73.23 & 82.35 & 83.00 & 78.66 & 80.77 \\
TCP \citep{Yao_2024_CVPR}
    & 82.63 & 78.20 & 80.35 & 82.77 & 58.07 & 68.25 & 91.63 & 74.73 & 82.32 & 87.13 & 80.77 & 83.83 \\
MMA \citep{seputis2024multi}
    & 82.27 & 78.57 & 80.38 & 83.20 & 65.63 & 73.38 & 85.46 & \textbf{82.34} & 83.87 & 86.23 & 80.03 & 82.20 \\
CoPrompt \citep{roy2023consistency}
    & 82.63 & 80.03 & 81.31 & 83.13 & 64.73 & 72.79 & 94.60 & 78.57 & 85.84 & 86.90 & 79.57 & 83.07 \\
HiCroPL \citep{Zheng_2025_ICCV}
    & \textbf{83.23} & 79.92 & 81.54 & 85.07 & 67.34 & 75.17 & \textbf{96.29} & 80.36 & \textbf{87.61} & 87.95 & \textbf{80.91} & \textbf{84.28} \\
MMRL \citep{guo2025mmrl}
    & 83.20 & 79.30 & 81.20 & \textbf{85.67} & 65.00 & 73.82 & 95.60 & 80.17 & 87.21 & \textbf{88.10} & 80.07 & 83.89 \\
PromptSRC \citep{khattak2023self}
    & 82.67 & 78.47 & 80.52 & 83.37 & 62.97 & 71.75 & 92.90 & 73.90 & 82.32 & 87.10 & 78.80 & 82.74 \\
\midrule
\rowcolor{avgblue}
\textbf{GAPL (Ours)}
    & 83.17 & \textbf{80.50} & \textbf{81.81} & 84.30 & \textbf{69.40} & \textbf{76.13} & 95.60 & 79.90 & 87.05 & 86.50 & 80.40 & 83.34 \\
\bottomrule
\end{tabular}%
}
\end{table*}

\subsection{Experimental Setup}

We implement GAPL in PyTorch with a CLIP ViT-B/16 backbone based on DVLP~\citep{Khattak_2023_CVPR, khattak2023self, zhang2024dept}. To preserve the pre-trained visual manifold, learnable prompts are inserted only into the text branch, with context length 4 and prompt depth 12, while the visual branch keeps both context length and depth at 0. We optimize the model with SGD. For base-to-novel generalization, we train for 5 epochs with batch size 4; for domain generalization, 20 epochs with batch size 16; and for cross-dataset transfer, batch size 32. The Contextual Modulator uses 4 learnable local signals and a hierarchical pooling scale of $M_i=i\times10$ following GalLoP~\citep{lafon2024gallop}. We set $\lambda_{txt}=25$ and $\lambda_{img}=10$, and follow the standard CoCoOp~\citep{zhou2022conditional} protocol for other training settings.

\subsection{Main Results}
\label{sec:main_results}

\noindent \textbf{Domain Generalization.}
We evaluate domain generalization by training on ImageNet and directly testing on its four variants. As shown in Table~\ref{tab:domain_generalization}, GAPL achieves 72.90\% on the source domain and the best average accuracy of 61.12\% across the four out-of-distribution datasets. In particular, GAPL improves over MMRL by 0.87\% on ImageNet, over TCP by 1.83\% on ImageNet-V2, and over PromptSRC by 0.58\% on ImageNet-Sketch. Although GAPL is not the top-performing method on every individual variant, it remains competitive on the more challenging datasets, reaching 50.43\% on ImageNet-A and 77.50\% on ImageNet-R. Overall, these results suggest that GAPL provides a favorable balance between source-domain performance and cross-domain robustness.

\noindent \textbf{Base-to-Novel Generalization.}
We further evaluate GAPL under the base-to-novel generalization protocol, where the model is trained on base classes and tested on unseen novel classes across 11 datasets. As reported in Table~\ref{tab:base_to_new}, GAPL achieves the best average Novel accuracy of 78.12\% and the best average harmonic mean (HM) of 81.77\%, while maintaining a competitive average Base accuracy of 85.78\%. In particular, GAPL obtains the highest HM on Caltech101, StanfordCars, Flowers102, Food101, SUN397, and DTD, and also achieves the best Novel accuracy on Flowers102, SUN397, and DTD. These results suggest that GAPL improves generalization to unseen classes while preserving strong performance on seen classes, leading to a favorable trade-off between base-class fitting and novel-class transfer.

\subsection{Ablation Study}
\label{ablation_study}

\noindent \textbf{Effectiveness of Components.}
We conduct a component-wise ablation to examine the roles of the three streams, as reported in Table~\ref{tab:ablation_components}. Using only the Global Invariant Stream ($\mathcal{G}$) gives an average accuracy of 58.77\%, which serves as the baseline of our framework. Adding the Gram-Anchored Stream ($\mathcal{S}$) improves the average accuracy to 59.81\%, with the most evident gain on ImageNet-A, where the accuracy increases from 47.10\% to 50.25\%. This suggests that the Gram-based second-order cue is particularly helpful under larger domain shifts. Adding the Contextual-Anchored Stream ($\mathcal{C}$) instead improves the average accuracy to 60.12\%, and yields stronger results on the source domain and ImageNet-R, indicating that local contextual alignment is beneficial for preserving discriminative visual-text correspondence. When all three streams are combined, the full GAPL model achieves the best overall average accuracy of 61.12\%. Compared with the two-branch variants, the full model consistently improves performance across the four target datasets, reaching 66.43\% on ImageNet-V2, 50.13\% on ImageNet-Sketch, 50.43\% on ImageNet-A, and 77.50\% on ImageNet-R. These results support the complementary roles of the three streams: $\mathcal{G}$ provides a stable semantic base, $\mathcal{S}$ improves robustness to domain variation through a Gram-based second-order cue, and $\mathcal{C}$ enhances local alignment. Their combination leads to the strongest overall trade-off between source-domain performance and cross-domain generalization.

\noindent \textbf{Impact of Visual Prompting Depth.}
We further examine the impact of visual prompting depth on cross-domain robustness by comparing our default configuration (\textbf{V0}) with a comprehensive deep prompting variant (\textbf{V12}), where learnable tokens are injected into all transformer layers of the visual encoder. As detailed in Table~\ref{tab:ablation_visual_depth}, while V12 achieves a marginally superior source accuracy of 73.00\%, it suffers a catastrophic degradation in out-of-distribution (OOD) robustness, with ImageNet-A performance plummeting to 47.30\%. This phenomenon underscores that deep visual prompting, despite enhancing in-distribution alignment, tends to distort the pre-trained feature manifold and leads to over-parameterization on the source domain, thereby compromising the universal representations of the frozen backbone. In contrast, our \textbf{V0} strategy maintains the integrity of the intrinsic visual priors and concentrates cross-modal adaptation within the final embedding space. This approach effectively safeguards the model's generalization capacity, ensuring superior stability and performance across diverse distribution shifts.

\noindent \textbf{Ablation on Second-Order Statistical Design.}
To examine whether richer second-order statistics are necessary, we compare three variants in the Gram-Anchored Stream: \emph{(i)} the diagonal-only Gram representation used in our final model, \emph{(ii)} a diagonal-plus-variance variant, and \emph{(iii)} the full Gram matrix. As shown in Table~\ref{tab:ablation_second_order_design}, directly using the full Gram matrix is infeasible in our setting due to excessive GPU memory consumption, resulting in out-of-memory errors. Meanwhile, augmenting the diagonal representation with additional variance statistics brings no performance gain: it yields the same source-domain accuracy (72.90\%) and slightly lower cross-domain average accuracy (61.10\% vs.\ 61.12\%) compared with the diagonal-only design. These results suggest that the useful statistical cue for prompt modulation is already captured by the compact diagonal descriptor, while introducing extra statistical complexity does not improve generalization. Therefore, following Occam's razor, we adopt the simplest effective design, namely the diagonal-only formulation.

\begin{table}[t]
\centering
\caption{Ablation on the statistical design of the Gram-Anchored Stream.}
\label{tab:ablation_second_order_design}
\small
\setlength{\tabcolsep}{10pt}
\begin{tabular}{lcc}
\toprule
\multirow{2}{*}{Statistic Design} & Source & Cross-domain \\
\cmidrule(lr){2-2} \cmidrule(lr){3-3}
& ImageNet & Average \\
\midrule
Diagonal only      & \textbf{72.90} & \textbf{61.12} \\
Diagonal + Variance & 72.90          & 61.10         \\
Full Gram Matrix   & \multicolumn{2}{c}{OOM} \\
\bottomrule
\end{tabular}
\end{table}

We make two key observations. First, the full Gram matrix is computationally prohibitive in our setting and leads to out-of-memory errors, indicating that directly modeling all pairwise channel correlations is impractical for our framework. Second, augmenting the diagonal representation with variance statistics brings no measurable performance gain over the diagonal-only design. This suggests that the main benefit of our method already lies in a compact channel-wise second-order descriptor, and that introducing extra statistical complexity does not further improve adaptation.

Therefore, following the principle of Occam's razor, we adopt the simplest effective design, namely the diagonal-only Gram representation. It achieves the same performance as the more elaborate diagonal-plus-variance variant, while remaining substantially more memory-efficient and implementation-friendly than the full Gram alternative.

\noindent \textbf{Sensitivity Analysis of Fusion Weight $\alpha$.}
We further investigate the sensitivity of the fusion weight $\alpha$ (defined in Eq.~\ref{eq:fusion}), which modulates the balance between the learned task-specific features $\mathbf{w}_{learn,c}$ and the generalized fixed priors $\mathbf{w}_{fixed,c}$.
The experimental results across a wide range of $\alpha \in [0.1, 0.9]$ are summarized in Table~\ref{tab:ablation_alpha}.
As $\alpha$ increases from $0.1$ to $0.7$, we observe a consistent improvement in out-of-distribution (OOD) performance, with the average accuracy rising from 59.38\% to a peak of 61.12\%.
Notably, the performance starts to saturate and exhibits a slight decline at $\alpha=0.9$ (60.96\% average).
This bell-shaped performance curve provides a crucial insight: while our learned prompts capture essential domain-specific discriminative cues, the fixed ensemble $w_{fixed,c}$ acts as an indispensable robustness anchor that prevents the model from over-parameterizing on the source domain.
The optimal performance achieved at $\alpha=0.7$ suggests that a high-fidelity fusion—leaning towards learned prompts while retaining a significant portion of frozen visual-textual priors—reaches the best trade-off for robust cross-domain generalization.
Consequently, we adopt $\alpha=0.7$ as our final configuration.

\begin{table}[h]
\centering
\caption{Sensitivity analysis of the fusion weight $\alpha$ in the Global Invariant Stream.}
\label{tab:ablation_alpha}
\small
\setlength{\tabcolsep}{8pt}
\begin{tabular}{l cccc c}
    \toprule
    $\alpha$ & -V2 & -Sketch & -A & -R & \textbf{Average} \\
    \midrule
    0.1 & 62.40 & 49.00 & 49.00 & 77.10 & 59.38 \\
    0.3 & 64.23 & 49.77 & 49.70 & 77.50 & 60.30 \\
    0.5 & 65.85 & 50.13 & 50.30 & \textbf{77.53} & 60.95 \\
    \rowcolor{avgblue}
    \textbf{0.7 (Final)} & \textbf{66.43} & \textbf{50.13} & \textbf{50.43} & 77.50 & \textbf{61.12} \\
    0.9 & \textbf{66.43} & 49.90 & 50.23 & 77.27 & 60.96 \\
    \bottomrule
\end{tabular}
\end{table}

\begin{table}[!t]
\centering
\caption{Ablation study of different stream combinations. $\mathcal{G}$: Global Invariant, $\mathcal{S}$: Gram-Anchored, $\mathcal{C}$: Contextual-Anchored.}
\label{tab:ablation_components}
\small
\setlength{\tabcolsep}{3.5pt}
\begin{tabular}{ccc c cccc c}
    \toprule
    $\mathcal{G}$ & $\mathcal{S}$ & $\mathcal{C}$ & Source & \multicolumn{4}{c}{Target (OOD Variants)} & \textbf{Average} \\
    \cmidrule(lr){5-8}
    & & & ImageNet & -V2 & -Sketch & -A & -R & \\
    \midrule
    \checkmark & & & 70.80 & 63.00 & 48.90 & 47.10 & 76.06 & 58.77 \\
    \checkmark & \checkmark & & 70.60 & 63.70 & 49.20 & 50.25 & 76.10 & 59.81 \\
    \checkmark & & \checkmark & \textbf{72.90} & 65.53 & 49.40 & 48.10 & 77.45 & 60.12 \\
    \midrule
    \rowcolor{avgblue}
    \checkmark & \checkmark & \checkmark & \textbf{72.90} & \textbf{66.43} & \textbf{50.13} & \textbf{50.43} & \textbf{77.50} & \textbf{61.12} \\
    \bottomrule
\end{tabular}
\end{table}

\begin{table}[!t]
\centering
\caption{Comparison of Visual Prompting Depth. \textbf{V0} (Ours) results are synchronized with final config.}
\label{tab:ablation_visual_depth}
\small
\setlength{\tabcolsep}{8pt}
\begin{tabular}{l c cccc}
    \toprule
    Setting & Source & \multicolumn{4}{c}{Target (OOD Variants)} \\
    \cmidrule(lr){3-6}
     & ImageNet & -V2 & -Sketch & -A & -R \\
    \midrule
    \rowcolor{avgblue}
    \textbf{V0 (Ours)} & 72.90 & \textbf{66.43} & \textbf{50.13} & \textbf{50.43} & \textbf{77.50} \\
    V12 (Deep) & \textbf{73.00} & 65.50 & 48.60 & 47.30 & 74.60 \\
    \bottomrule
\end{tabular}
\end{table}

\subsection{Further Analysis}
\label{sec:further_analysis}

\begin{figure}[t]
  \centering
  \includegraphics[width=\columnwidth]{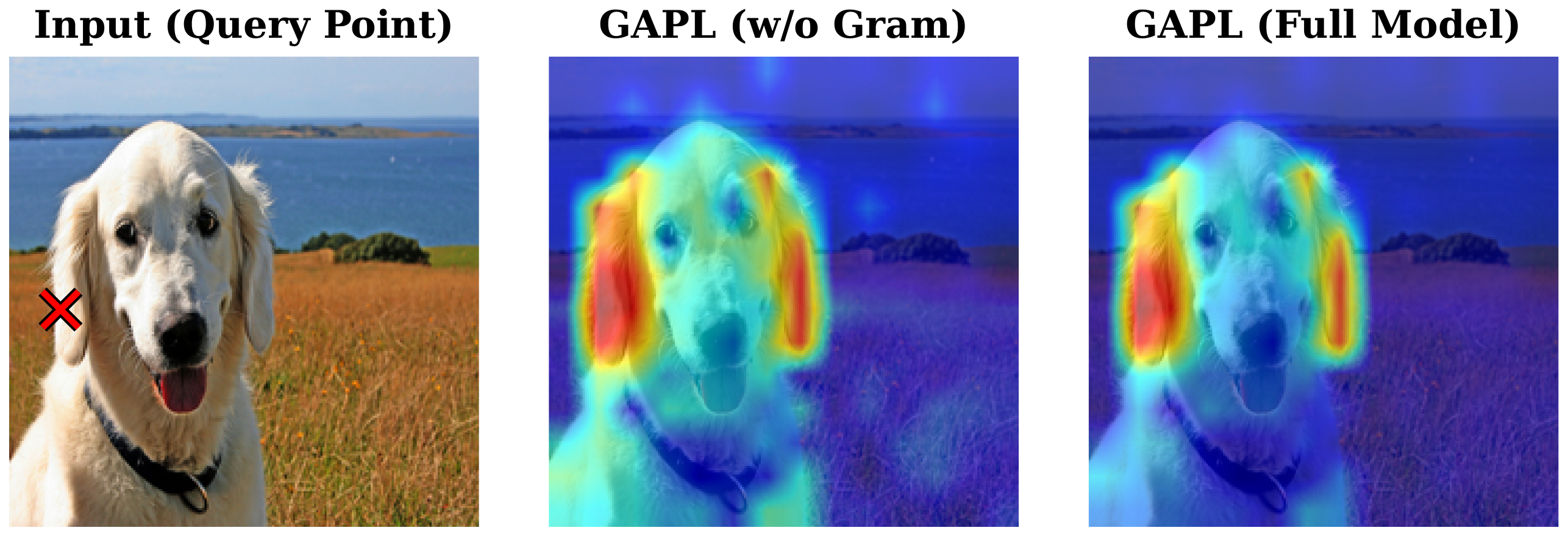}
  \caption{Visual specificity comparison (query point on the dog's ear).}

  \label{fig:visualization}
\end{figure}

\noindent \textbf{Semantic Specificity Analysis.}
To better understand the effect of the Gram-Anchored Stream, we visualize patch-level cosine similarity maps. We choose a query point (marked by a red cross) on a semantic part---the dog's ear---and compute its similarity to all other patches. As shown in Figure~\ref{fig:visualization}, the variant without the Gram-Anchored Stream produces a relatively diffuse response: although the ear region is activated, high responses also extend to nearby body regions. In contrast, the full GAPL model yields a more localized similarity map, with stronger responses concentrated around the ear itself. This comparison suggests that the Gram-based anchor helps suppress irrelevant patch responses and improves semantic specificity at the part level.

\begin{figure}[t]
  \centering
  \includegraphics[width=\columnwidth]{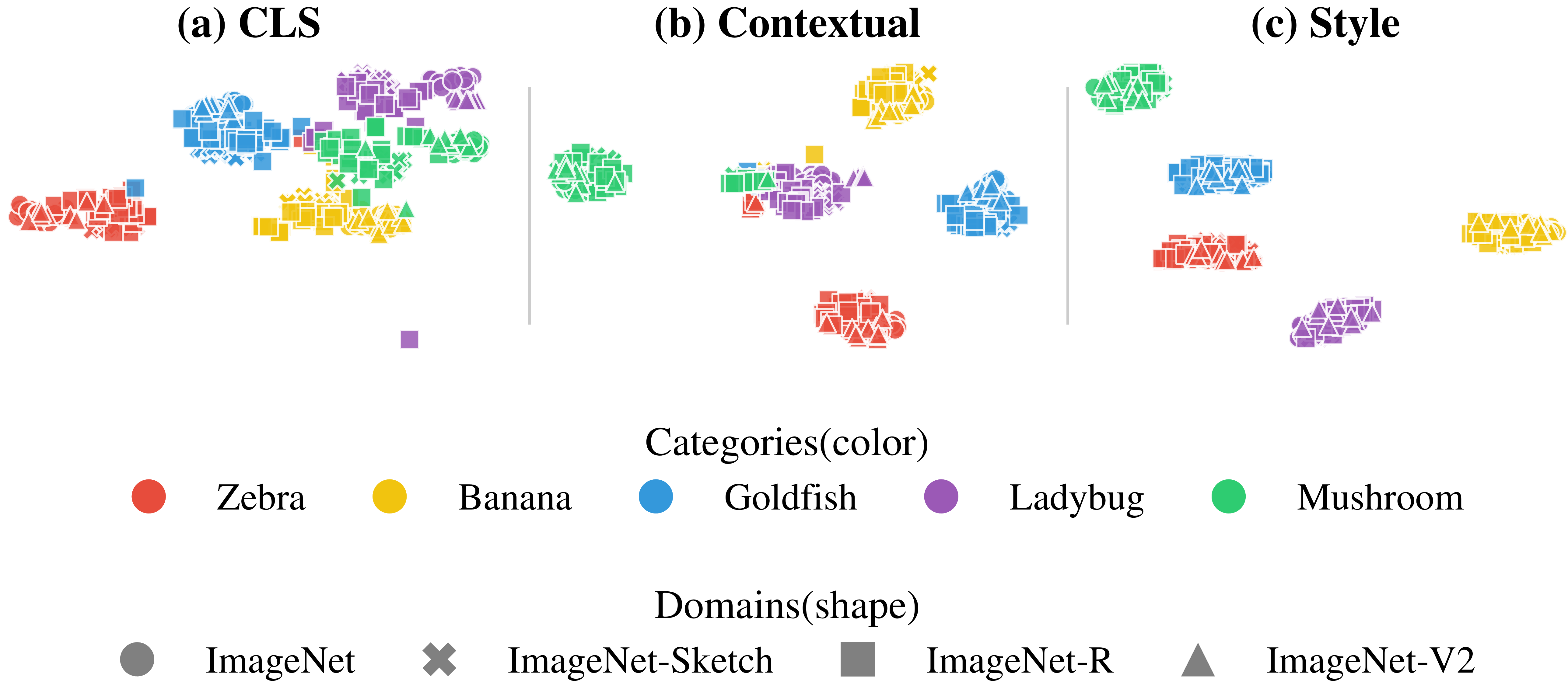}
  \caption{t-SNE visualization of latent manifolds across four domains for five selected classes.
  \textbf{(a) CLS Token} (first-order global) exhibits severe domain divergence with scattered distributions.
  \textbf{(b) Contextual Text Anchor} (first-order local) remains susceptible to stylistic noise despite local patch aggregation.
  \textbf{(c) Style Text Anchor} (ours, second-order) achieves superior alignment by collapsing domain-specific variance into compact, class-discriminative clusters.}
  \label{fig:tsne}
\end{figure}

\begin{figure}[t]
  \centering
  \includegraphics[width=\columnwidth]{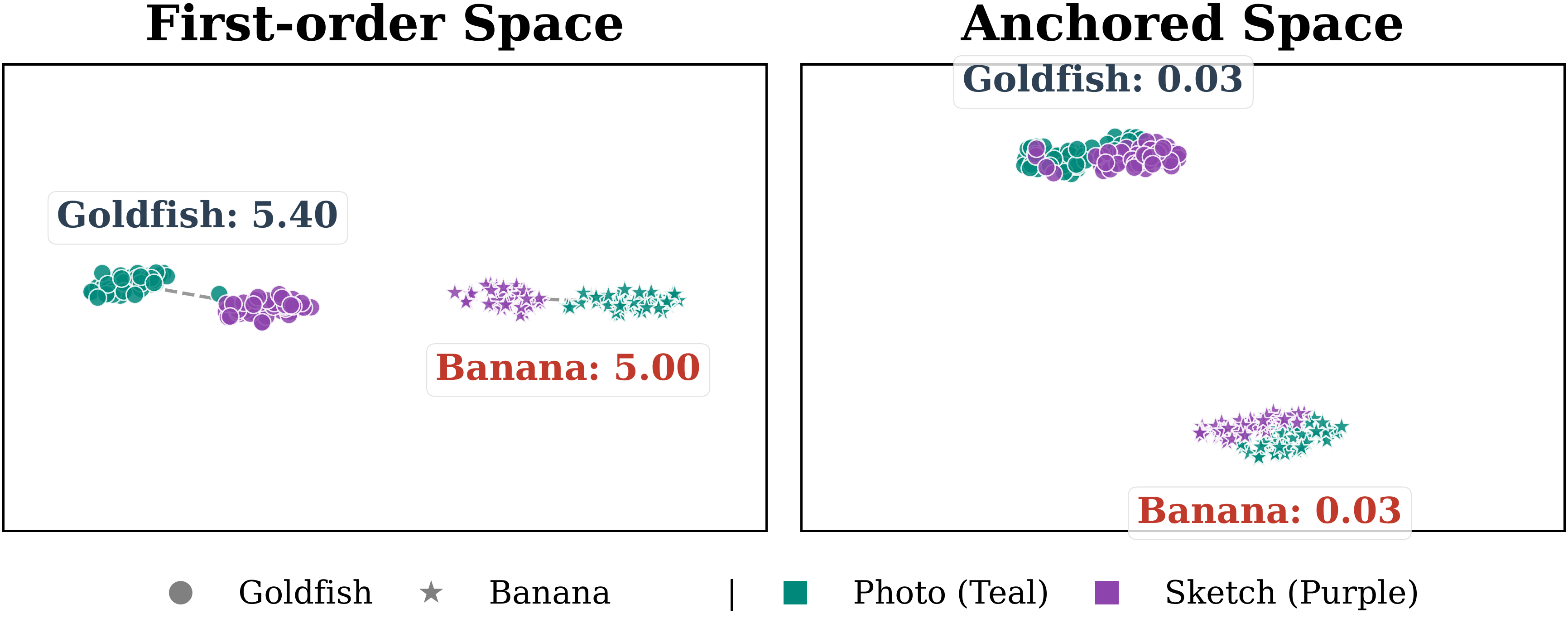}
  \caption{Quantitative analysis of domain alignment. We visualize the feature distribution of Photo (Teal) and Sketch (Purple) domains for ``Goldfish'' and ``Banana''. }
  \label{fig:domain_gap}
\end{figure}

\noindent \textbf{Latent Manifold Homogeneity.}
We further visualize the feature distribution with t-SNE~\citep{maaten2008visualizing} to examine how different representations behave across domains. In Figure~\ref{fig:tsne}(a), the standard CLS token features show noticeable domain separation and relatively scattered intra-class distributions. Figure~\ref{fig:tsne}(b) shows that the Contextual Text Anchor improves semantic grouping to some extent, but samples from different domains are still not well aligned. In comparison, Figure~\ref{fig:tsne}(c) shows that the Style Text Anchor produces more compact class-wise clusters and better cross-domain mixing. These visualizations provide qualitative evidence that the Gram-based second-order cue helps reduce domain-induced variation in the feature space.

\noindent \textbf{Quantitative Domain Alignment.}
To complement the visualization results, we further measure the alignment between the ``Photo'' and ``Sketch'' domains for two representative classes. As shown in Figure~\ref{fig:domain_gap}, we compute the Euclidean distance between the domain centroids in the feature space. In the first-order space, the centroid distances are relatively large (5.40 for Goldfish and 5.00 for Banana), indicating clear domain separation. In the anchored space, the corresponding distances decrease substantially to 0.03 for both classes. Although this analysis is limited to representative examples, it is consistent with the qualitative results above and suggests that the proposed anchored representation can better align samples from different domains.

\section{Conclusion}
\label{sec:conclusion}
In this paper, we presented \textbf{Gram-Anchored Prompt Learning (GAPL)}, a prompt learning framework that improves vision-language model adaptation by introducing a Gram-based second-order visual cue. Instead of relying only on first-order visual conditioning, GAPL uses a Gram-based Style Modulator to adjust prompted text representations with image-level statistical information. Experiments on 15 datasets show that GAPL achieves a favorable trade-off between source-domain performance and cross-domain generalization. Additional analyses further suggest that the proposed Gram-based anchor helps improve semantic specificity and cross-domain feature alignment. 

\noindent \textbf{Limitations and future work.}
Our current framework still performs inference with fixed parameters after offline training. A natural next step is to extend GAPL to test-time adaptation or continual domain generalization, so that the model can better handle dynamically changing distributions. A second direction is to explore whether the proposed Gram-based anchoring strategy can benefit dense prediction tasks, such as zero-shot image segmentation, where both semantic alignment and spatial localization are important.






\bibliographystyle{IEEEtranN}
\bibliography{sample-base}

\end{document}